\documentclass{article}
\usepackage{ijcai25}

\usepackage{times}
\usepackage{soul}
\usepackage{url}
\usepackage[utf8]{inputenc}
\usepackage[small]{caption}
\usepackage{graphicx}
\usepackage{amsmath}
\usepackage{amsthm}
\usepackage{booktabs}
\usepackage{algorithm}
\usepackage{algorithmic}
\usepackage[switch]{lineno}
\usepackage{multirow}
\usepackage{xcolor}
\usepackage{amssymb} 
\usepackage{marvosym}

\title{The Role of Video Generation in Enhancing Data-Limited Action Understanding}

\author{Wei Li$^{1,3}$ \and
 Dezhao Luo$^{4}$ \and
 Dongbao Yang$^{1}$ \textsuperscript{\Letter} \and
 Zhenhang Li$^{1,3}$ \\
 Weiping Wang$^{1}$ \and
 Yu Zhou $^{2}$ \textsuperscript{\Letter}
 \affiliations
$^1$Institute of Information Engineering, Chinese Academy of Sciences\\
$^2$ VCIP \& TMCC \& DISSec, College of Computer Science, Nankai University \\
$^3$ School of Cyber Security, University of Chinese Academy of Sciences\\
$^4$Queen Mary University of London
\emails
\{liwei1,wangweiping,lizhenhang, yangdongbao\}@iie.ac.cn, yzhou@nankai.edu.cn, 
dezhao.luo@qmul.ac.uk}

\begin{document}

\maketitle

\begin{abstract}

Video action understanding tasks in real-world scenarios always suffer data limitations. In this paper, we address the data-limited action understanding problem by bridging data scarcity. We propose a novel method that employs a text-to-video diffusion transformer to generate annotated data for model training. This paradigm enables the generation of realistic annotated data on an infinite scale without human intervention. We proposed the information enhancement strategy and the uncertainty-based label smoothing tailored to generate sample training. Through quantitative and qualitative analysis, we observed that real samples generally contain a richer level of information than generated samples. Based on this observation, the information enhancement strategy is proposed to enhance the informative content of the generated samples from two aspects: the environments and the characters. Furthermore, we observed that some low-quality generated samples might negatively affect model training. To address this, we devised the uncertainty-based label smoothing strategy to increase the smoothing of these samples, thus reducing their impact. We demonstrate the effectiveness of the proposed method on four datasets across five tasks and achieve state-of-the-art performance for zero-shot action recognition.

\end{abstract}

\section{Introduction}
\let\thefootnote\relax
\footnotetext{{\Letter} Corresponding author.}

\label{sec:intro}

\begin{figure}[ht]
    \flushleft
    \includegraphics[width=1\columnwidth]{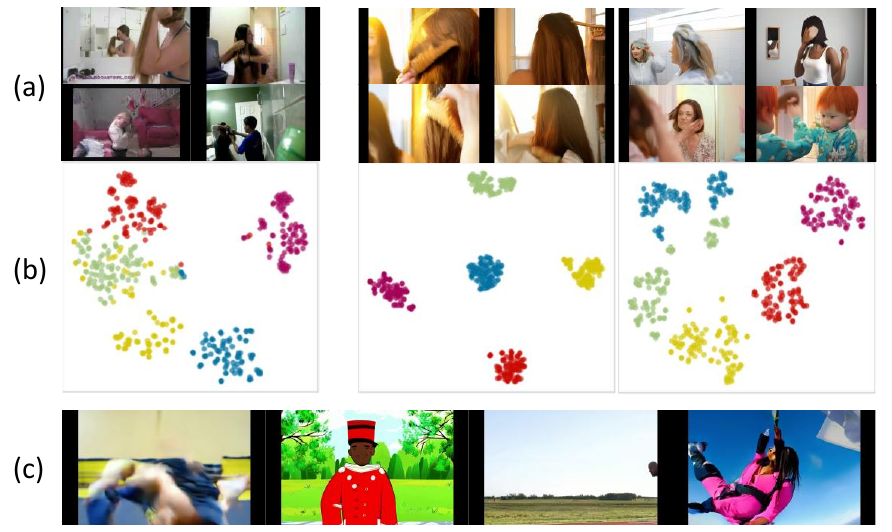}
     \caption{
     The samples (a) and t-SNE visualizations (b) of the synthetic dataset and the real dataset. From left to right, they are: the HMDB-51 dataset, the synthetic HMDB-51 dataset with the basic strategy and the synthetic HMDB-51 dataset with our proposed information enhancement strategy. (c) Unsatisfactory synthetic videos.}
     \label{fig:clust}
 \end{figure}

Over the past decade, deep learning ~\cite{lecun2015deep,miech2019howto100m} has brought remarkable progress benefiting from large-scale annotated data~\cite{carreira2018short,sultani2018real}, especially in the computer vision community such as CLIP with 400M image-text pairs~\cite{radford2021learning}. However, obtaining high-quality datasets is often complex, time-consuming, and costly~\cite{carreira2018short,grauman2022ego4d}, requiring a large amount of manual annotation. In particular in the field of video understanding, human annotations can be inaccurate (50\% samples are not aligned in the HowTo100M dataset ~\cite{miech2019howto100m}). A lack of a well-annotated dataset with sufficient samples will limit the learning of models and result in poor generalization ability, facing challenges in real-world applications such as abnormal action detection~\cite{sultani2018real,lv2023unbiased} and long-tail action recognition~\cite{perrett2023use,grauman2022ego4d}.

In order to solve the data-limited problem in video understanding, previous methods often focus on designing data augmentation strategies ~\cite{yun2020videomix,kim2020learning,li2022video} and transfer knowledge ~\cite{kim2025leveraging,li2024large,rasheed2023fine,luo2023towards} from other modalities, such as images, where data is easier to collect. However, we argue that these solutions are suboptimal as a) data augmentation does not create novel semantics for general knowledge learning~\cite{trabuccoeffective}, and b) knowledge learned from other modalities does not provide necessary information for video understanding, specifically, images do not include temporal relations between video frames.

With the development of diffusion models ~\cite{yang2024cogvideox,liu2024sora,esser2023structure}, existing methods propose to synthesis datasets to improve model learning~\cite{he2023synthetic,tian2024stablerep,luo2024generative,feng2023diverse}. 
There is a problem in the utilization of generated samples for training video understanding models, manifested by the deficient information within the samples, which leads to suboptimal training efficiency. As illustrated in Figure~\ref{fig:clust} (a) and (b), our finding is that due to the complex and rich information of real videos, real samples have higher intra-class distances and lower inter-class distances than synthetic samples; while the videos generated by the diffusion transformer tend to include information with simple and similar content, which limits the learnable information contained in the synthetic samples .  This also makes the classification difficulty of synthetic samples lower than that of real samples.

To solve the problem of deficient information within generated samples, we propose an information enhancement strategy to enhance the effective information of synthetic samples, which refers to context information conducive to action understanding. This strategy injects various character and environment information that conforms to specific human actions into the generated video and alleviates the domain gap between synthetic and real samples. Due to the limited capabilities of the text2video model, some synthetic videos perform unsatisfactorily and contain less effective semantic information, as shown in Figure~\ref{fig:clust} (c). To alleviate the impact of low-quality synthetic samples on model training, we propose uncertainty-based label smoothing. We calculate the uncertainty of the generated samples with CLIP to measure their quality and adjust the smoothness of label smoothing with uncertainty, which can prevent the model from overfitting low-quality samples and alleviating its impact.

We conducted extensive experiments on four datasets (Kinetics-600, UCF-101, HMDB-51, UCF-Crime) across five tasks (few-shot, zero-shot, base-to-novel, long-tail, abnormal action detection) and demonstrated the effectiveness of our method in data-limited action understanding. We achieved state-of-the-art performance for zero-shot action recognition tasks on Kinetics-600, UCF-101, and HMDB-51.

In summary, our contributions are as follows:

\textbf{(1)} To the best of our knowledge, we are the first to investigate the effectiveness of data generated by video diffusion transformers for enhanced data-limited video understanding that includes action recognition and action detection tasks.

\textbf{(2)}  To solve the problem of deficient information within generated samples, we propose two strategies: the information enhancement strategy and the uncertainty-based label smoothing. These two strategies significantly improve the training efficiency of the generated samples.

\textbf{(3)} We conduct extensive experiments on four datasets across five data-limited action understanding tasks to demonstrate the effectiveness of our proposed method, and our method achieves SOTA performance for zero-shot action recognition.

\section{Related Work}
\label{sec:relatedworks}

\subsection{Training with Synthetic Samples}
The methods of training with synthetic samples can be broadly categorized into two main streams: generative-based and graphics engine-based methods. The graphics engine-based methods use real objects' 3D models to arrange, reconstruct, or move them according to certain rules and then render them to generate training samples. It has found usage in a myriad of fields, including object detection~\cite{gaidon2016virtual,cabon2020virtual}, optical flow estimation~\cite{dosovitskiy2015flownet,kim2022transferable}, auto driving~\cite{dosovitskiy2017carla}, etc.

With the explosion development of AIGC, image generation models have become of sufficient quality to generate training samples. Earlier works on image understanding have concentrated on the generation of samples or data augmentation with GANs ~\cite{baranchuk2021label,li2022bigdatasetgan}. But with the emergence of the diffusion model, many methods focus on generating samples using diffusion models given its realistic, high-quality and controllable generative ability.~\cite{he2023synthetic} found that diffusion model synthetic samples can significantly improve the classification results of zero-shot, few-shot and transfer learning, and designed strategies to filter the noise of synthetic samples and enhance their diversity.~\cite{feng2023diverse} proposes a novel  test-time prompt tuning method which leverages diffusion models to generate augmented data.~\cite{zhou2023training} proposes diffusion inversion, which invert images to the diffusion model's latent space, and generates novel samples by conditioning the generative model.~\cite{feng2024instagen} fine-tunes a diffusion model to enhance the capability of the object detection model.~\cite{wanggenerated} studies the reasons why synthetic samples sometimes harms contrastive learning performance from the perspective of data inflation and data augmentation, and proposes adaptive inflation to improve the contrastive learning.~\cite{trabuccoeffective} uses a diffusion model to edit and change the semantics of the image to address the lack of diversity in previous image augmentation methods.

Within the realm of video understanding, most works still rely on graphics engines or GANs~\cite{guo2022learning,kim2022transferable}. Unlike these previous works, this paper focuses on leveraging text2video diffusion transformers to improve data-limited action understanding that includes action recognition and action detection tasks. To the best of our knowledge, no prior research has delved into this domain.

\subsection{Text-to-Video Diffusion Models}
In recent years, text-to-video generation has undergone significant advancements, particularly driven by the rapid development of diffusion models. Previous video generation models based on diffusion typically followed a UNet-based architecture. Video Diffusion Models (VDMs)~\cite{ho2022video} extend image synthesis diffusion models to video generation by training jointly on both image and video data. Gen-1~\cite{esser2023structure} presents a structure and content-aware model that modifies videos guided by example images or texts. With the impressive capabilities demonstrated by Sora~\cite{liu2024sora}, which uses Transformer as the backbone of diffusion models, i.e. Diffusion Transformers (DiT), has gradually become mainstream. A series of DiT-based works have emerged, including CogVideoX~\cite{yang2024cogvideox} and Hunyuan Video~\cite{kong2024hunyuanvideo}. These works are capable of generating high-resolution videos with coherent actions, which makes it possible to apply the generated long-tail video data for downstream task training. In this paper, we utilize CogVideoX~\cite{yang2024cogvideox} to generate all training samples because it performs well in generating human-related videos and includes rich motion information.

\section{Method}
\label{sec: Method}

\begin{figure*}
    \flushleft
    \centering
    \includegraphics[width=2\columnwidth]{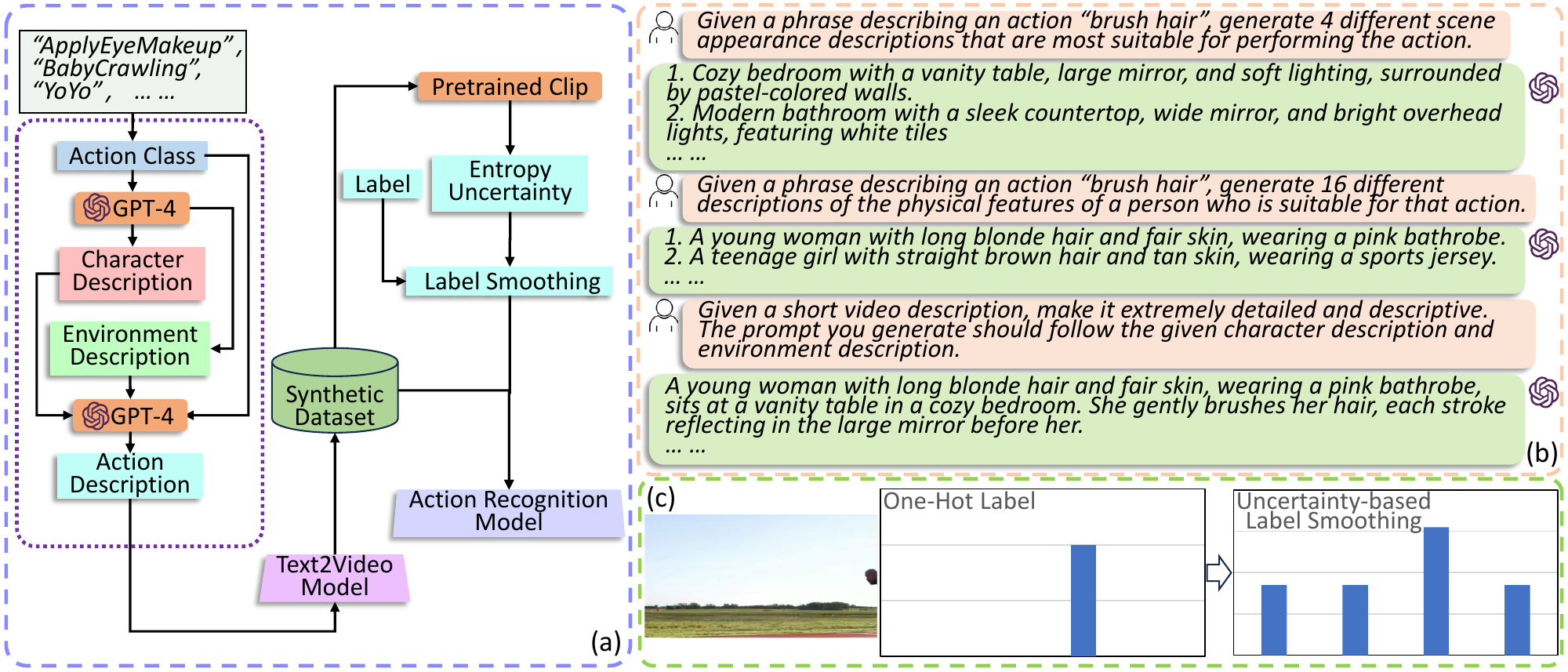}
     \caption{(a): The overall structure of our proposed method. We designed two strategies for generating sample training, information enhancement strategy (left) and uncertainty-based label smoothing strategy (right). (b): The process of generating action description information through proposed information enhancement strategy. (c): Uncertainty-based label smoothing uses a higher smoothness for low-quality generated samples with higher uncertainty.}
     \label{fig:overall}
 \end{figure*}
 
\subsection{Overview}
We address the data-limited action understanding problem by bridging the data scarcity. To this end, we adopt the text2video model to generate the required data according to the target task. Our approach consists of two steps: 1. Generate a video dataset with a text2video model based on given labels; 2. Train the action understanding model with the generated video dataset, as shown in Figure~\ref{fig:overall} (a). This paradigm is concise yet general and can produce infinite-scale annotated data without human intervention.

\subsection{Video Sample Generator}
We employ the text2video model to solve the problem of data-limited action understanding by generating synthetic samples. Given a data-limited 
dataset $D=\{v_1,v_2,\dots,v_{n-1},v_n\}$ with a class name list $C=\{c_1,c_2,\dots,c_{k-1},c_k\}$, we use GPT-4o to generate text descriptions of humans performing specific actions, then generate the synthetic video dataset $D'$  based on them:

\begin{equation}
D'=DiT(gpt(prompt_{act},c))
\end{equation}
where $DiT$ represents the text2video diffusion transformer. 

Although this paradigm is straightforward and simple, it has the problem of insufficient information in the generated samples, which leads to suboptimal training efficiency. When action descriptions are generated solely on the basis of action names, text-to-video models tend to produce videos with simplistic and repetitive content with these descriptions, leading to lower information richness in the generated videos compared to real videos.

\subsection{Information Enhancement Strategy}

To enrich the action-related information that serves as a crucial factor in action understanding datasets to train action understanding models effectively, we propose the information Enhancement Strategy towards generated samples. Action-relevant information, such as the environment in which the action frequently occurs, the objects that often appear when the action is performed, the characteristics of the people who always perform the action, benefits action understanding model training, as it encompasses rich semantics and contextual details that are relevant to human actions. A well-constructed generated dataset should encompass sufficient action-related information, as this determines the upper limit of knowledge that the model can acquire from it.

Given an action category name $c$, we construct the contextual information most relevant to this action. We mainly focus on information from two perspectives: the environment where the action occurs and the appearance characteristics of the person performing the action. We generate contextual information through GPT-4o, where we generate 4 different environment descriptions $Env= gpt(prompt_{env},c)$ and 16 different character descriptions $Char=gpt(prompt_{char},c)$ for each action. Finally, we use this contextual information to enhance the content of the generated video. The process of generating action description information is shown in Figure~\ref{fig:overall} (b).

\begin{equation}
\begin{split}
D'=DiT(gpt(Char,Env,prompt_{act},c))
\end{split}
\end{equation}

\subsection{Uncertainty-Based Label Smoothing}
Due to the limited capabilities of the text2video model, generated videos sometimes have suboptimal generation effects and contain less effective semantic information for training. Such samples may be detrimental to the training of the model, as shown in Figure~\ref{fig:clust}. 

To address this problem, we propose uncertainty-based label smoothing. Given a synthetic video sample $v'$, we first use CLIP~\cite{radford2021learning} to calculate its similarity $S$ with all category names $C$, and then use the entropy of the similarity to measure its uncertainty.
\begin{equation}
\begin{split}
S=\lbrace CLIP(v',c)|c \in C\rbrace,\\
H(S)=-\sum_{i=1}^{k}S_ilog\,S_i
\end{split}
\end{equation}
After that, we dynamically adjust the smoothing of label smoothing according to the uncertainty of the sample:
\begin{equation}
q'_i= \begin{cases}
1-\varepsilon,\,\,\,\,\,\,\,\,\,\,\,\,\,if\,i=y\\
\varepsilon/(K-1),otherwise
\end{cases},\varepsilon=wH(S)
\end{equation}
whle $w$ represents the weighting of uncertainty-based label smoothing, $y=[q_1,q_2,...,q_{k-1},q_{k}]$ represents the original one-hot label. As low-quality generated samples often exhibit greater uncertainty, leading to an increased degree of label smoothing in such cases, as shown in Figure~\ref{fig:overall} (c). In this way, we can prevent the model from overfitting to low-quality samples and mitigate their impact on model training.

\section{Experiment}
\label{sec:Experiment}

\subsection{Implementation Details}
The text2video model we use is CogVideoX-2B~\cite{yang2024cogvideox}. For each dataset, we generate 128 videos for each category with 50 inference steps. We need to emphasize that the training data of CogVideoX-2B does not contain the annotated UCF-101, HMDB-51 or Kinetics-600 datasets we used in our experiments, so there is no data leakage. For information enhancement strategy,
we use GPT-4o to generate multiple different environment descriptions at once, with the following prompt like \textit{``Given a phrase describing an action, generate four different scene appearance descriptions that are most suitable for performing the action.''}. Similarly, the prompt for generating character descriptions is as follows: \textit{``Given a phrase describing an action, generate 16 different descriptions of the physical features of a person who is suitable for that action.''}. In subsequent training, we adopt TC-CLIP~\cite{kim2025leveraging} in most experiments except for abnormal action detection. For abnormal action detection, we use the X-CLIP-B/32~\cite{ni2022expanding} following~\cite{lv2023unbiased}. For tasks where real samples are not available such as zero-shot, we train the model with synthetic samples only. For tasks where real samples are available such as few-shot, long-tail, etc., we pre-train the model with synthetic samples and then fine-tune with the real samples. We set the $w$ to 0.3 in uncertainty-based label smoothing. After calculating the uncertainty of all samples in a dataset, normalization is performed on them to constrain the scale.

\begin{table*}[htb]
\centering
\begin{tabular}{lccccc}
\toprule
Method       & HMDB-51      & UCF-101    & K600 (Top-1) & K600 (Top-5) & All (Top-1)\\
\midrule
Vanilla CLIP~\cite{radford2021learning} & 40.8 ± 0.3 & 63.2 ± 0.2 & 59.8 ± 0.3 & 83.5 ± 0.2 & 54.6 \\
ActionCLIP~\cite{wang2021actionclip} & 49.1 ± 0.4 & 68.0 ± 0.9 & 56.1 ± 0.9 & 83.2 ± 0.2 & 57.7 \\
X-CLIP~\cite{ni2022expanding} & 44.6 ± 5.2 & 72.0 ± 2.3 & 65.2 ± 0.4 & 86.1 ± 0.8 & 60.6 \\
Vita-CLIP~\cite{wasim2023vita} & 48.6 ± 0.6 & 75.0 ± 0.6 & 67.4 ± 0.5 & - & 63.7 \\
Open-VCLIP~\cite{weng2023open}  & 53.9 ± 1.2 & 83.4 ± 1.2 & 73.0 ± 0.8 & 93.2 ± 0.1 & 70.1 \\
OTI ~\cite{zhu2023orthogonal} &54.2 ± 1.3  &83.3 ± 0.3 &66.9 ± 1.0 & - & 68.1 \\
OST~\cite{chen2024ost} & 55.9 ± 1.2 & 79.7 ± 1.1 & 75.1 ± 0.6 & 94.6 ± 0.2 & 70.2 \\
FROSTER~\cite{huangfroster}  & 54.8 ± 1.3 & 84.8 ± 1.1 & 74.8 ± 0.9 & - & 71.5 \\
\midrule
IMP-MoE-L (VIT-L) ~\cite{akbari2023alternating} &59.1  &91.5 & 76.8&  -&  75.8\\

OTI (VIT-L) ~\cite{zhu2023orthogonal} &59.3 ± 1.7  &88.1 ± 1.0 &70.6 ± 0.5 & - & 72.7 \\
\midrule
ViFi-CLIP~\cite{rasheed2023fine}& 51.3 ± 0.6 & 76.8 ± 0.7 & 71.2 ± 1.0 & 92.2 ± 0.3 & 66.4 \\
ViFi-CLIP~\cite{rasheed2023fine}+Ours&60.8 ± 0.4  &  81.9 ± 1.2&  76.0 ± 1.0&  94.3 ± 0.2& 72.9 \\
\hline
BIKE~\cite{Wu_2023_CVPR} &53.3 ± 1.1  & 79.6 ± 0.3 &68.5 ± 1.3  & 90.9 ± 0.4 &  67.1\\
BIKE~\cite{Wu_2023_CVPR}+Ours& 56.5 ± 0.7& 87.7 ± 1.0 & 78.0 ± 1.1 & 94.7 ± 0.3& 74.1 \\
\hline
TC-CLIP~\cite{kim2025leveraging} & 56.0 ± 0.3 & 85.4 ± 0.8 & 78.1 ± 1.0 & 95.7 ± 0.3 & 73.2 \\
TC-CLIP~\cite{kim2025leveraging}+Ours& 61.0 ± 1.0 & 86.5 ± 0.7& 78.6 ± 0.8 &96.1 ± 0.2  & 75.3 \\
\bottomrule
\end{tabular}
 \caption{Comparison with state-of-the-arts on zero-shot action recognition.}
 \label{zero-shot1}
\end{table*}

\begin{table*}[htb]
\centering
\label{tab:few_shot_action_recognition_comparison}
\begin{tabular}{l c c c c|c c c c c}
\toprule
\multirow{2}{*}{Method} & \multicolumn{4}{c}{HMDB-51} & \multicolumn{4}{c}{UCF-101} \\

&K=2&K=4&K=8&K=16&K=2&K=4&K=8&K=16\\
\midrule
Vanilla CLIP~\cite{radford2021learning} & 41.9 & 41.9 & 41.9 & 41.9 & 63.6 & 63.6 & 63.6 & 63.6  \\
ActionCLIP~\cite{wang2021actionclip} & 47.5 & 57.9 & 57.3 & 59.1 & 70.6 & 71.5 & 73.0 & 91.4   \\
X-CLIP~\cite{ni2022expanding} & 53.0 & 57.3 & 62.8 & 64.0 & 76.4 & 83.4 & 88.3 & 91.4   \\
ViFi-CLIP~\cite{rasheed2023fine} & 57.2 & 62.7 & 64.5 & 66.8 & 80.7 & 85.1 & 90.0 & 92.7  \\
OST~\cite{chen2024ost} & 59.1 & 62.9 & 64.9 & 68.2 & 82.5 & 87.5 & 91.7 & 93.9   \\
\hline
TC-CLIP~\cite{kim2025leveraging}  & 58.6 & 63.3 & 65.5 & 68.8 & 86.8 & 90.1 & 92.0 & 94.3   \\
TC-CLIP~\cite{kim2025leveraging}+Ours & 63.0 & 65.6 & 68.9 & 71.4 & 88.1 & 91.3 & 92.9 & 94.0  \\
\bottomrule
\end{tabular}
\caption{Comparison with state-of-the-arts on few-shot action recognition. All the models are directly fine-tuned from CLIP.}
 \label{few-shot1}
\end{table*}

\begin{table*}[htb]
\centering
\label{tab:base_to_novel_generalization}
\begin{tabular}{l c c c|c c c}
\toprule
\multirow{2}{*}{Method} & \multicolumn{3}{c}{HMDB-51} & \multicolumn{3}{c}{UCF-101} \\

&Base&Novel&HM&Base&Novel&HM\\
\midrule
Vanilla CLIP~\cite{radford2021learning}  & 53.3 & 46.8 & 49.8 & 78.5 & 63.6 & 70.3 \\
ActionCLIP~\cite{wang2021actionclip}  & 69.1 & 37.3 & 48.5 & 90.1 & 58.1 & 70.7 \\
X-CLIP~\cite{ni2022expanding}  & 69.4 & 45.5 & 55.0 & 89.9 & 58.9 & 71.2 \\
ViFi-CLIP~\cite{rasheed2023fine}  & 73.8 & 53.3 & 61.9 & 92.9 & 67.7 & 78.3  \\
Open-VCLIP~\cite{weng2023open} & 70.3 & 50.4 & 58.9 & 94.8 & 77.5 & 85.3 \\
FROSTER~\cite{huangfroster} & 74.1 & 58.0 & 65.1 & 95.3 & 80.0 & 87.0 \\
\hline
TC-CLIP~\cite{kim2025leveraging}  & 73.3 & 59.1 & 65.5 & 95.4 & 81.6 & 88.0 \\
TC-CLIP~\cite{kim2025leveraging}+Ours  & 77.7 & 64.0 & 70.2 & 95.5 &  85.2 &  90.1\\
\bottomrule
\end{tabular}
\caption{Comparison with state-of-the-arts on base-to-novel generalization. All the models are directly fine-tuned from CLIP.}
\label{base-novel1}
\end{table*}

\begin{table*}[htb]
\centering
\label{tab:Longtail Action Recognition}
\begin{tabular}{l c c c c |c c c c }
\toprule
\multirow{2}{*}{Method} & \multicolumn{4}{c}{HMDB-51} & \multicolumn{4}{c}{UCF-101} \\
&Few&Tail&Head&Acc&Few&Tail&Head&Acc\\
\midrule
TC-CLIP~\cite{kim2025leveraging}  & 49.41 & 79.63 & 87.08 & 60.39 & 76.13 & 66.48  &  80.03& 73.46   \\
TC-CLIP~\cite{kim2025leveraging}+Ours  & 52.55 & 80.37 & 87.50 & 62.40 & 79.30 & 68.36&  86.25 & 78.77   \\
\bottomrule
\end{tabular}
\caption{Results on long-tail action recognition.}
\label{long-tail1}
\end{table*}


\begin{table}[htb]
\centering

\label{tab:anomaly_detection_comparison}
\begin{tabular}{lcc}
\toprule
Method & AUC& AUC\textsubscript{A} \\
\midrule
SVM Baseline & 50.00 & 50.00 \\
Sohrab et al.~\cite{sohrab2018subspace} & 58.50 & - \\
BODS~\cite{wang2019gods} & 68.26 & - \\
GODS~\cite{wang2019gods} & 70.46 & - \\
\hline
Zhang et al.~\cite{zhang2019temporal} & 78.66 & - \\
Motion-Aware~\cite{zhu2019motion} & 79.10 & 62.18 \\
Wu et al.~\cite{wu2020not} & 82.44 & - \\

\hline
MIL & 81.80 & 59.90 \\
MIL+Ours & 84.03 & 62.44 \\
\bottomrule
\end{tabular}
\caption{Results on abnormal action detection.}
\label{abnormal-detection1}
\end{table}

\begin{table}[htb]
\centering
\begin{tabular}{lc}
\toprule
Method       & HMDB-51    \\
\midrule
TC-CLIP~\cite{kim2025leveraging} &  56.0\\
+Basic&  57.6\\
+Cha &  60.0\\
+Env &  59.7\\
+IE &  61.0\\

\bottomrule
\end{tabular}
 \caption{Ablation study about information enhancement strategy. ``IE'' refers to proposed information enhancement strategy with character information and environment information.}
 \label{ablation1}
\end{table}

\begin{table}[htb]
\centering
\begin{tabular}{lcc}
\toprule
Method       &UCF-101& HMDB-51    \\
\midrule
TC-CLIP~\cite{kim2025leveraging} &85.4&  56.0\\
+LS &85.8&60.7\\
+UW  & 86.0&60.4 \\
+UF  & 86.5&60.3\\
+UL  & 86.5&61.0 \\
\bottomrule
\end{tabular}
 \caption{Ablation study about uncertainty-based label smoothing.}
 \label{ablation2}
\end{table}

\subsection{Main Results}
\subsubsection{Zero-shot Action Recognition}
In zero-shot action recognition, only the category names of the target dataset are available. The model needs to perform action recognition without training on target dataset. We conduct zero-shot action recognition experiments on the UCF-101, HMDB-51, and Kinetics-600 datasets. We report the Top-1 accuracy for UCF-101 and HMDB-51, Top-1 and Top-5 accuracy for Kinetics-600. 

Table~\ref{zero-shot1} shows the zero-shot action recognition results of the proposed method. The proposed method has significant improvements across all three datasets, especially the HMDB-51, where the Top-1 accuracy of zero-shot action recognition is improved by 5\% after combining the generated sample training. It should be noted that when our method is based solely on the VIT-B backbone, our performance on HMDB-51 even surpasses those of methods based on the VIT-L backbone (~\cite{zhu2023orthogonal,akbari2023alternating}). The rich contextual information related to actions is introduced by the proposed method, leading to the improvement of zero-shot action recognition.

The proposed method is generic and can be adapted to various models. We train using the generated samples on three zero-shot action recognition models, including ViFi-CLIP, BIKE, and TC-CLIP. The results are shown in Table~\ref{zero-shot1} bottom. The proposed method improves all three models, especially the Top-1 accuracy of BIKE on Kinetics-600 is improved by almost 10\%.

\subsubsection{Few-shot Action Recognition}
The purpose of the few-shot action recognition task is to enable the model to accurately classify action categories using only a few labeled samples per category, thereby alleviating the reliance on data. We conducted the few-shot action recognition experiments with the UCF-101 and HMDB-51 datasets in Table~\ref{few-shot1}. We first pre-train the model with generated samples and then fine-tune it on each dataset with only K samples per category, where K is in 2, 4, 8 and 16. 

The results demonstrate that the proposed method achieves improvements in all K-Shot settings,  which can be attributed to the effective initialization of the model with the proposed method. Table~\ref{few-shot1} shows that the improvement gradually decreases as the K increases from 2 to 16. This phenomenon could be attributed to the fact that the information provided by the generated samples partially overlaps with that provided by the real samples. The proposed information enhancement strategy enhances the action-related information in the generated samples, allowing the proposed method to remain effective even at high values of K=16 in HMDB-51 and K=8 in UCF-101.

\subsubsection{Base-to-Novel Generalization}
The task of base-to-novel generalization is proposed to evaluate the generalization of a model to unseen classes when only samples from half of the classes are available for training. In each dataset, 16 samples per category are selected from half of the categories to construct the base split for training, while the remaining half of the categories serve as the novel split for evaluation.  

Table~\ref{base-novel1} shows the Top-1 accuracy of the base classes, the novel classes, and their harmonic mean (HM) in UCF-101 and HMDB-51. The proposed method shows noticeable gains for base and novel categories.
\subsubsection{Long-Tail Action Recognition}
The concept of ``long-tail problem'' refers to the phenomenon where the imbalance in the distribution of samples across different classes leads to a significantly larger number of samples in the minority head classes compared to the tail classes, as models tend to perform well on the head classes but exhibit poorer performance on the tail classes. Following~\cite{perrett2023use}, we construct long-tail action recognition datasets based on UCF-101 and HMDB-51. 

The results of long-tail action recognition are presented in Table~\ref{long-tail1}. The proposed method can improve the performance of the tail and the few categories while maintaining the performance of the head categories.

\subsubsection{Abnormal Action Detection}
Video anomaly action detection aims to identify abnormal events or actions within videos. The primary objective is to pinpoint the specific time window when an anomalous activity occurs, such as crimes, traffic accidents, or other illegal activities, etc. This task is challenging because weak labels are provided only at the video level, whereas the model needs to make frame-level predictions for abnormal actions. We employ the MIL (Multiple Instance Learning) baseline method following~\cite{lv2023unbiased} for abnormal action detection. Initially, we first pre-train the model on generated samples and subsequently fine-tune it on real samples. We report the AUC and AUC$_A$~\cite{lv2021localizing} of the frame-level ROC results on the UCF-Crime dataset.

The results for abnormal action detection are presented in Table~\ref{abnormal-detection1}. The proposed method is effective for abnormal action detection, leading to a 2.23\% improvement in AUC and a 2.54\% improvement in AUC$_A$ compared to MIL baseline. The collection of real-world abnormal action videos is often constrained by real conditions, such as the rarity of the occurrence of anomalous actions. The proposed methods enable the low-cost construction of anomalous action datasets, which effectively enhances the model's performance.

\subsection{Ablation Studies}

\begin{figure}[ht]
    \flushleft
    \includegraphics[width=1\columnwidth]{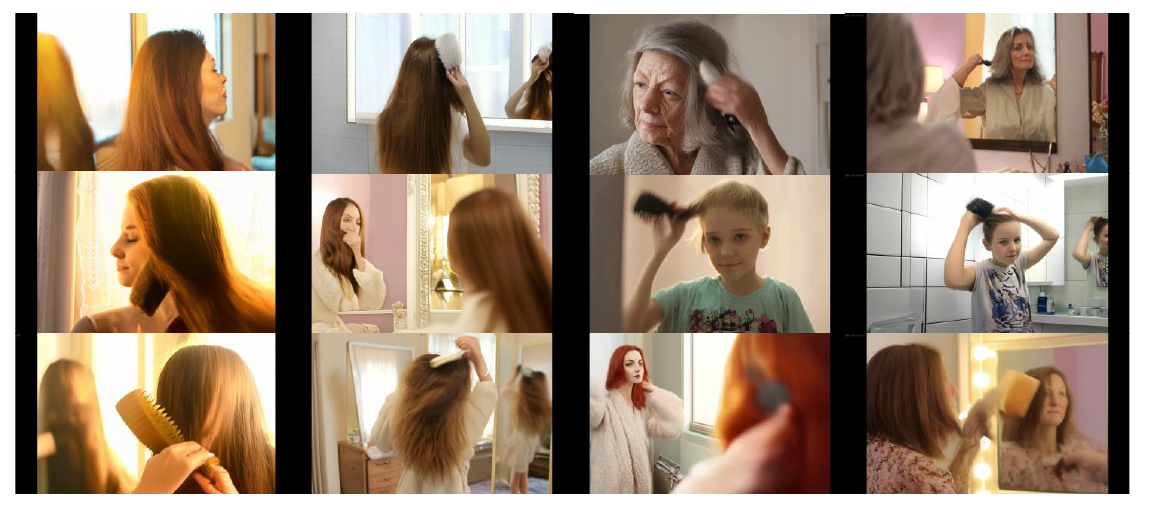}
     \caption{Visualization of generated samples. The strategies adopted from left to right are: ``Basic'', ``Env'', ``Cha'' and ``IE''.}
     \label{fig:IEsamples}
 \end{figure}

We discuss the effectiveness of the proposed strategies in Table~\ref{ablation1} and Table~\ref{ablation2}. We report the Top-1 accuracy under zero-shot setting for for the ablation study.

The proposed information enhancement strategy (``IE'') consists of two components: environmental information enhancement and character information enhancement. We discuss them in Table~\ref{ablation1}. The ``Basic'' strategy refers to generating action descriptions solely based on category names and then using them to create videos; The ``Env'' strategy incorporates environmental information where the action occurs; the ``Cha'' strategy incorporates character appearance information about the person performing the action. 

The results show that incorporating either environmental information or character information in the synthetic samples improves zero-shot accuracy. The information enhancement strategy, which combines environmental information and character information, can lead to a 3.4\% improvement compared to the basic strategy. This indicates that enhancing the action-related information of the generated samples is beneficial for training action understanding models. Figure~\ref{fig:IEsamples} illustrates the visualization of the ``brush hair'' action samples generated using the four aforementioned strategies. When employing the basic strategy, the video generation model tends to generate similar videos in which the characters consistently appear as Caucasian women with long brown hair. When the ``Env'' strategy is adopted, various contextual details that assist in understanding the "brush hair" action are included in generated videos, such as full-length mirrors and bathrooms with tiled backgrounds. This information enhances the knowledge encompassed by the generated dataset. We set the information enhancement strategy as the default setting for the remaining experiments.

In Table~\ref{ablation2}, we discuss three strategies to mitigate the impact of low-quality generated samples on training. ``LS'' refers to the label smoothing without additional design toward low-quality generated samples. ``UW'' denotes the uncertainty-based weighting strategy, where we weight the loss based on the uncertainty of each sample. ``UF'' denotes the uncertainty-based filtering strategy that removes high-uncertainty samples from the synthetic dataset.  ``UL'' denotes the uncertainty-based label smoothing strategy, where we adjust the smoothness of the label smoothing based on the uncertainty of each sample.

Based on the experiments, uncertainty-based label smoothing shows the best performance among three strategies, leading to a 1.1\% improvement on the UCF-101 dataset. This validates the effectiveness of the proposed uncertainty-based label smoothing strategy, which can prevent the model from overfitting low-quality samples and alleviating its impact.

\section{Analysis }

\begin{table}[ht]
\centering
\begin{tabular}{lcccc}
\toprule
Method  &DB& MMD\textsubscript{$lin$}&MMD\textsubscript{$brf$}&MD\textsubscript{$poly$} \\
\midrule
Basic       &2.120 &0.115  &0.098  &0.130  \\
IE          &4.310&0.112  &0.092  &0.122  \\
HMDB-51     &5.040&0.000&0.000&0.000\\
UCF-101     &2.840&0.036  &0.031  &0.042  \\
\bottomrule
\end{tabular}
 \caption{Davies Bouldin scores (DB) and Maximum Mean Discrepancy (MMD) of real and synthetic datasets. ``IE'' refers to generated HMDB-51 dataset with information enhancement strategy, ``Basic'' refers to generated HMDB-51 dataset with basic strategy. We list the MMD values between various datasets and the HMDB-51 dataset.}
\label{analysis2}
\end{table}

We conducted qualitative analysis on synthetic datasets and real datasets, as shown in Figure~\ref{fig:clust}. We extract CLIP features from the HMDB-51 dataset, the synthetic HMDB-51 dataset with the basic strategy, and the synthetic HMDB-51 dataset with the information enhancement strategy. Subsequently, we make the t-SNE visualization of these datasets based on the extracted features.

In Figure~\ref{fig:clust}, it is evident that the boundaries between different categories of synthetic samples are quite distinct. In contrast, the boundaries between samples of different categories in the real dataset are more blurred and lack clarity. This could be attributed to the higher complexity of real samples compared to synthetic samples. This complexity stems from various aspects such as changes in lighting, camera angles, and background details, among others, leading real samples to contain more information. When generating samples based on the basic strategy, the text2video model tends to produce videos with similar content, resulting in low levels of effective information contained in the generated datasets. This limits the upper bound of knowledge that the synthetic data can contain.

The proposed information enhancement strategy effectively increases the information content in the generated samples. As illustrated in Figure~\ref{fig:clust}, the proposed method improves the information contained in the generated samples, bringing their distribution patterns closer to real samples.

We perform quantitative analyzes in Table~\ref{analysis2}. Davies-Bouldin (DB) is a metric used to assess the effectiveness of clustering algorithms, where a smaller value indicates clearer separation between clusters and tighter intra-cluster cohesion. We employ the Davies-Bouldin score to measure the complexity of the datasets. The first column of Table~\ref{analysis2} presents the Davies-Bouldin scores for the generated datasets and real datasets. Real datasets (UCF-101,HMDB-51) exhibit higher DB scores compared to generated samples due to their complexity. The Maximum Mean Discrepancy(MMD)~\cite{gretton2006kernel} is a commonly used metric for measuring the domain gap between different datasets, where a higher MMD indicates a larger domain gap. In columns 2, 3, and 4 of Table~\ref{analysis2}, we present the MMD between various datasets and the HMDB-51 dataset. MMD between real datasets, such as UCF-101 and HMDB-51, exhibit lower values, while the MMD between generated datasets and real datasets are higher, indicating a larger domain gap. The proposed strategy reduces the domain gap between generated samples and real samples, which is manifested as lower MMD values.

\section{Conclusion}
In this paper, we tackle the challenge of data-limited action understanding by introducing a generic method that leverages synthetic video data generated through the text2video diffusion transformer. This approach enables the cost-effective creation of large-scale annotated video datasets, significantly mitigating the data scarcity in action understanding tasks. To enhance the utility of generated samples, we propose the information enhancement strategy and uncertainty-based label smoothing. We validate the proposed method on four datasets across five tasks and achieve SOTA performance for zero-shot action recognition.

\section*{Acknowledgements}
Supported by the National Natural Science Foundation of China (Grant NO 62376266 and 62406318), Key Laboratory of Ethnic Language Intelligent Analysis and Security Governance of MOE, Minzu University of China. 

\label{sec:formatting}

\bibliographystyle{named}
\bibliography{ijcai25}

\begin{thebibliography}{}

\bibitem[\protect\citeauthoryear{Akbari \bgroup \em et al.\egroup }{2023}]{akbari2023alternating}
Hassan Akbari, Dan Kondratyuk, Yin Cui, Rachel Hornung, Huisheng Wang, and Hartwig Adam.
\newblock Alternating gradient descent and mixture-of-experts for integrated multimodal perception.
\newblock {\em NeurIPS}, 36:79142--79154, 2023.

\bibitem[\protect\citeauthoryear{Baranchuk \bgroup \em et al.\egroup }{2021}]{baranchuk2021label}
Dmitry Baranchuk, Ivan Rubachev, Andrey Voynov, Valentin Khrulkov, and Artem Babenko.
\newblock Label-efficient semantic segmentation with diffusion models.
\newblock {\em arXiv preprint arXiv:2112.03126}, 2021.

\bibitem[\protect\citeauthoryear{Cabon \bgroup \em et al.\egroup }{2020}]{cabon2020virtual}
Yohann Cabon, Naila Murray, and Martin Humenberger.
\newblock Virtual kitti 2.
\newblock {\em arXiv preprint arXiv:2001.10773}, 2020.

\bibitem[\protect\citeauthoryear{Carreira \bgroup \em et al.\egroup }{2018}]{carreira2018short}
Joao Carreira, Eric Noland, Andras Banki-Horvath, Chloe Hillier, and Andrew Zisserman.
\newblock A short note about kinetics-600.
\newblock {\em arXiv preprint arXiv:1808.01340}, 2018.

\bibitem[\protect\citeauthoryear{Chen \bgroup \em et al.\egroup }{2024}]{chen2024ost}
Tongjia Chen, Hongshan Yu, Zhengeng Yang, Zechuan Li, Wei Sun, and Chen Chen.
\newblock Ost: Refining text knowledge with optimal spatio-temporal descriptor for general video recognition.
\newblock In {\em CVPR}, pages 18888--18898, 2024.

\bibitem[\protect\citeauthoryear{Dosovitskiy \bgroup \em et al.\egroup }{2015}]{dosovitskiy2015flownet}
Alexey Dosovitskiy, Philipp Fischer, Eddy Ilg, Philip Hausser, Caner Hazirbas, Vladimir Golkov, Patrick Van Der~Smagt, Daniel Cremers, and Thomas Brox.
\newblock Flownet: Learning optical flow with convolutional networks.
\newblock In {\em ICCV}, pages 2758--2766, 2015.

\bibitem[\protect\citeauthoryear{Dosovitskiy \bgroup \em et al.\egroup }{2017}]{dosovitskiy2017carla}
Alexey Dosovitskiy, German Ros, Felipe Codevilla, Antonio Lopez, and Vladlen Koltun.
\newblock Carla: An open urban driving simulator.
\newblock In {\em Conference on robot learning}, pages 1--16. PMLR, 2017.

\bibitem[\protect\citeauthoryear{Esser \bgroup \em et al.\egroup }{2023}]{esser2023structure}
Patrick Esser, Johnathan Chiu, Parmida Atighehchian, Jonathan Granskog, and Anastasis Germanidis.
\newblock Structure and content-guided video synthesis with diffusion models.
\newblock In {\em ICCV}, pages 7346--7356, 2023.

\bibitem[\protect\citeauthoryear{Feng \bgroup \em et al.\egroup }{2023}]{feng2023diverse}
Chun-Mei Feng, Kai Yu, Yong Liu, Salman Khan, and Wangmeng Zuo.
\newblock Diverse data augmentation with diffusions for effective test-time prompt tuning.
\newblock In {\em ICCV}, pages 2704--2714, 2023.

\bibitem[\protect\citeauthoryear{Feng \bgroup \em et al.\egroup }{2024}]{feng2024instagen}
Chengjian Feng, Yujie Zhong, Zequn Jie, Weidi Xie, and Lin Ma.
\newblock Instagen: Enhancing object detection by training on synthetic dataset.
\newblock In {\em CVPR}, pages 14121--14130, 2024.

\bibitem[\protect\citeauthoryear{Gaidon \bgroup \em et al.\egroup }{2016}]{gaidon2016virtual}
Adrien Gaidon, Qiao Wang, Yohann Cabon, and Eleonora Vig.
\newblock Virtual worlds as proxy for multi-object tracking analysis.
\newblock In {\em CVPR}, pages 4340--4349, 2016.

\bibitem[\protect\citeauthoryear{Grauman \bgroup \em et al.\egroup }{2022}]{grauman2022ego4d}
Kristen Grauman, Andrew Westbury, Eugene Byrne, Zachary Chavis, Antonino Furnari, Rohit Girdhar, Jackson Hamburger, Hao Jiang, Miao Liu, Xingyu Liu, et~al.
\newblock Ego4d: Around the world in 3,000 hours of egocentric video.
\newblock In {\em CVPR}, pages 18995--19012, 2022.

\bibitem[\protect\citeauthoryear{Gretton \bgroup \em et al.\egroup }{2006}]{gretton2006kernel}
Arthur Gretton, Karsten Borgwardt, Malte Rasch, Bernhard Sch{\"o}lkopf, and Alex Smola.
\newblock A kernel method for the two-sample-problem.
\newblock {\em NeurIPS}, 19, 2006.

\bibitem[\protect\citeauthoryear{Guo \bgroup \em et al.\egroup }{2022}]{guo2022learning}
Xi~Guo, Wei Wu, Dongliang Wang, Jing Su, Haisheng Su, Weihao Gan, Jian Huang, and Qin Yang.
\newblock Learning video representations of human motion from synthetic data.
\newblock In {\em CVPR}, pages 20197--20207, 2022.

\bibitem[\protect\citeauthoryear{He \bgroup \em et al.\egroup }{2023}]{he2023synthetic}
Ruifei He, Shuyang Sun, Xin Yu, Chuhui Xue, Wenqing Zhang, Philip Torr, Song Bai, and Xiaojuan Qi.
\newblock Is synthetic data from generative models ready for image recognition?
\newblock In {\em ICLR}, 2023.

\bibitem[\protect\citeauthoryear{Ho \bgroup \em et al.\egroup }{2022}]{ho2022video}
Jonathan Ho, Tim Salimans, Alexey Gritsenko, William Chan, Mohammad Norouzi, and David~J Fleet.
\newblock Video diffusion models.
\newblock {\em NeurIPS}, 35:8633--8646, 2022.

\bibitem[\protect\citeauthoryear{Huang \bgroup \em et al.\egroup }{2024}]{huangfroster}
Xiaohu Huang, Hao Zhou, Kun Yao, and Kai Han.
\newblock Froster: Frozen clip is a strong teacher for open-vocabulary action recognition.
\newblock In {\em ICLR}, 2024.

\bibitem[\protect\citeauthoryear{Kim \bgroup \em et al.\egroup }{2020}]{kim2020learning}
Taeoh Kim, Hyeongmin Lee, MyeongAh Cho, Ho~Seong Lee, Dong~Heon Cho, and Sangyoun Lee.
\newblock Learning temporally invariant and localizable features via data augmentation for video recognition.
\newblock In {\em ECCV}, pages 386--403. Springer, 2020.

\bibitem[\protect\citeauthoryear{Kim \bgroup \em et al.\egroup }{2022}]{kim2022transferable}
Yo-whan Kim, Samarth Mishra, SouYoung Jin, Rameswar Panda, Hilde Kuehne, Leonid Karlinsky, Venkatesh Saligrama, Kate Saenko, Aude Oliva, and Rogerio Feris.
\newblock How transferable are video representations based on synthetic data?
\newblock {\em NeurIPS}, 35:35710--35723, 2022.

\bibitem[\protect\citeauthoryear{Kim \bgroup \em et al.\egroup }{2025}]{kim2025leveraging}
Minji Kim, Dongyoon Han, Taekyung Kim, and Bohyung Han.
\newblock Leveraging temporal contextualization for video action recognition.
\newblock In {\em ECCV}, pages 74--91. Springer, 2025.

\bibitem[\protect\citeauthoryear{Kong \bgroup \em et al.\egroup }{2024}]{kong2024hunyuanvideo}
Weijie Kong, Qi~Tian, Zijian Zhang, Rox Min, Zuozhuo Dai, Jin Zhou, Jiangfeng Xiong, Xin Li, Bo~Wu, Jianwei Zhang, et~al.
\newblock Hunyuanvideo: A systematic framework for large video generative models.
\newblock {\em arXiv preprint arXiv:2412.03603}, 2024.

\bibitem[\protect\citeauthoryear{LeCun \bgroup \em et al.\egroup }{2015}]{lecun2015deep}
Yann LeCun, Yoshua Bengio, and Geoffrey Hinton.
\newblock Deep learning.
\newblock {\em nature}, 521(7553):436--444, 2015.

\bibitem[\protect\citeauthoryear{Li \bgroup \em et al.\egroup }{2022a}]{li2022bigdatasetgan}
Daiqing Li, Huan Ling, Seung~Wook Kim, Karsten Kreis, Sanja Fidler, and Antonio Torralba.
\newblock Bigdatasetgan: Synthesizing imagenet with pixel-wise annotations.
\newblock In {\em CVPR}, pages 21330--21340, 2022.

\bibitem[\protect\citeauthoryear{Li \bgroup \em et al.\egroup }{2022b}]{li2022video}
Wei Li, Dezhao Luo, Bo~Fang, Xiaoni Li, Yu~Zhou, and Weiping Wang.
\newblock Video motion perception for self-supervised representation learning.
\newblock In {\em International conference on artificial neural networks}, pages 508--520. Springer, 2022.

\bibitem[\protect\citeauthoryear{Li \bgroup \em et al.\egroup }{2024}]{li2024large}
Wei Li, Dezhao Luo, Dongbao Yang, and Weiping Wang.
\newblock Large language model for action anticipation.
\newblock In {\em International Conference on Artificial Neural Networks}, pages 207--222. Springer, 2024.

\bibitem[\protect\citeauthoryear{Liu \bgroup \em et al.\egroup }{2024}]{liu2024sora}
Yixin Liu, Kai Zhang, Yuan Li, Zhiling Yan, Chujie Gao, Ruoxi Chen, Zhengqing Yuan, Yue Huang, Hanchi Sun, Jianfeng Gao, et~al.
\newblock Sora: A review on background, technology, limitations, and opportunities of large vision models.
\newblock {\em arXiv preprint arXiv:2402.17177}, 2024.

\bibitem[\protect\citeauthoryear{Luo \bgroup \em et al.\egroup }{2023}]{luo2023towards}
Dezhao Luo, Jiabo Huang, Shaogang Gong, Hailin Jin, and Yang Liu.
\newblock Towards generalisable video moment retrieval: Visual-dynamic injection to image-text pre-training.
\newblock In {\em Proceedings of the IEEE/CVF Conference on Computer Vision and Pattern Recognition}, pages 23045--23055, 2023.

\bibitem[\protect\citeauthoryear{Luo \bgroup \em et al.\egroup }{2024}]{luo2024generative}
Dezhao Luo, Shaogang Gong, Jiabo Huang, Hailin Jin, and Yang Liu.
\newblock Generative video diffusion for unseen cross-domain video moment retrieval.
\newblock {\em CoRR}, 2024.

\bibitem[\protect\citeauthoryear{Lv \bgroup \em et al.\egroup }{2021}]{lv2021localizing}
Hui Lv, Chuanwei Zhou, Zhen Cui, Chunyan Xu, Yong Li, and Jian Yang.
\newblock Localizing anomalies from weakly-labeled videos.
\newblock {\em TIP}, 30:4505--4515, 2021.

\bibitem[\protect\citeauthoryear{Lv \bgroup \em et al.\egroup }{2023}]{lv2023unbiased}
Hui Lv, Zhongqi Yue, Qianru Sun, Bin Luo, Zhen Cui, and Hanwang Zhang.
\newblock Unbiased multiple instance learning for weakly supervised video anomaly detection.
\newblock In {\em CVPR}, pages 8022--8031, 2023.

\bibitem[\protect\citeauthoryear{Miech \bgroup \em et al.\egroup }{2019}]{miech2019howto100m}
Antoine Miech, Dimitri Zhukov, Jean-Baptiste Alayrac, Makarand Tapaswi, Ivan Laptev, and Josef Sivic.
\newblock Howto100m: Learning a text-video embedding by watching hundred million narrated video clips.
\newblock In {\em ICCV}, pages 2630--2640, 2019.

\bibitem[\protect\citeauthoryear{Ni \bgroup \em et al.\egroup }{2022}]{ni2022expanding}
Bolin Ni, Houwen Peng, Minghao Chen, Songyang Zhang, Gaofeng Meng, Jianlong Fu, Shiming Xiang, and Haibin Ling.
\newblock Expanding language-image pretrained models for general video recognition.
\newblock In {\em ECCV}, pages 1--18. Springer, 2022.

\bibitem[\protect\citeauthoryear{Perrett \bgroup \em et al.\egroup }{2023}]{perrett2023use}
Toby Perrett, Saptarshi Sinha, Tilo Burghardt, Majid Mirmehdi, and Dima Damen.
\newblock Use your head: Improving long-tail video recognition.
\newblock In {\em CVPR}, pages 2415--2425, 2023.

\bibitem[\protect\citeauthoryear{Radford \bgroup \em et al.\egroup }{2021}]{radford2021learning}
Alec Radford, Jong~Wook Kim, Chris Hallacy, Aditya Ramesh, Gabriel Goh, Sandhini Agarwal, Girish Sastry, Amanda Askell, Pamela Mishkin, Jack Clark, et~al.
\newblock Learning transferable visual models from natural language supervision.
\newblock In {\em ICML}, pages 8748--8763. PMLR, 2021.

\bibitem[\protect\citeauthoryear{Rasheed \bgroup \em et al.\egroup }{2023}]{rasheed2023fine}
Hanoona Rasheed, Muhammad~Uzair Khattak, Muhammad Maaz, Salman Khan, and Fahad~Shahbaz Khan.
\newblock Fine-tuned clip models are efficient video learners.
\newblock In {\em CVPR}, pages 6545--6554, 2023.

\bibitem[\protect\citeauthoryear{Sohrab \bgroup \em et al.\egroup }{2018}]{sohrab2018subspace}
Fahad Sohrab, Jenni Raitoharju, Moncef Gabbouj, and Alexandros Iosifidis.
\newblock Subspace support vector data description.
\newblock In {\em ICPR}, pages 722--727. IEEE, 2018.

\bibitem[\protect\citeauthoryear{Sultani \bgroup \em et al.\egroup }{2018}]{sultani2018real}
Waqas Sultani, Chen Chen, and Mubarak Shah.
\newblock Real-world anomaly detection in surveillance videos.
\newblock In {\em CVPR}, pages 6479--6488, 2018.

\bibitem[\protect\citeauthoryear{Tian \bgroup \em et al.\egroup }{2024}]{tian2024stablerep}
Yonglong Tian, Lijie Fan, Phillip Isola, Huiwen Chang, and Dilip Krishnan.
\newblock Stablerep: Synthetic images from text-to-image models make strong visual representation learners.
\newblock {\em NeurIPS}, 36, 2024.

\bibitem[\protect\citeauthoryear{Trabucco \bgroup \em et al.\egroup }{2024}]{trabuccoeffective}
Brandon Trabucco, Kyle Doherty, Max~A Gurinas, and Ruslan Salakhutdinov.
\newblock Effective data augmentation with diffusion models.
\newblock In {\em ICLR}, 2024.

\bibitem[\protect\citeauthoryear{Wang and Cherian}{2019}]{wang2019gods}
Jue Wang and Anoop Cherian.
\newblock Gods: Generalized one-class discriminative subspaces for anomaly detection.
\newblock In {\em ICCV}, pages 8201--8211, 2019.

\bibitem[\protect\citeauthoryear{Wang \bgroup \em et al.\egroup }{2021}]{wang2021actionclip}
Mengmeng Wang, Jiazheng Xing, and Yong Liu.
\newblock Actionclip: A new paradigm for video action recognition.
\newblock {\em arXiv preprint arXiv:2109.08472}, 2021.

\bibitem[\protect\citeauthoryear{Wang \bgroup \em et al.\egroup }{2024}]{wanggenerated}
Yifei Wang, Jizhe Zhang, and Yisen Wang.
\newblock Do generated data always help contrastive learning?
\newblock In {\em ICLR}, 2024.

\bibitem[\protect\citeauthoryear{Wasim \bgroup \em et al.\egroup }{2023}]{wasim2023vita}
Syed~Talal Wasim, Muzammal Naseer, Salman Khan, Fahad~Shahbaz Khan, and Mubarak Shah.
\newblock Vita-clip: Video and text adaptive clip via multimodal prompting.
\newblock In {\em CVPR}, pages 23034--23044, 2023.

\bibitem[\protect\citeauthoryear{Weng \bgroup \em et al.\egroup }{2023}]{weng2023open}
Zejia Weng, Xitong Yang, Ang Li, Zuxuan Wu, and Yu-Gang Jiang.
\newblock Open-vclip: Transforming clip to an open-vocabulary video model via interpolated weight optimization.
\newblock In {\em ICML}, pages 36978--36989. PMLR, 2023.

\bibitem[\protect\citeauthoryear{Wu \bgroup \em et al.\egroup }{2020}]{wu2020not}
Peng Wu, Jing Liu, Yujia Shi, Yujia Sun, Fangtao Shao, Zhaoyang Wu, and Zhiwei Yang.
\newblock Not only look, but also listen: Learning multimodal violence detection under weak supervision.
\newblock In {\em ECCV}, pages 322--339. Springer, 2020.

\bibitem[\protect\citeauthoryear{Wu \bgroup \em et al.\egroup }{2023}]{Wu_2023_CVPR}
Wenhao Wu, Xiaohan Wang, Haipeng Luo, Jingdong Wang, Yi~Yang, and Wanli Ouyang.
\newblock Bidirectional cross-modal knowledge exploration for video recognition with pre-trained vision-language models.
\newblock In {\em CVPR}, pages 6620--6630, June 2023.

\bibitem[\protect\citeauthoryear{Yang \bgroup \em et al.\egroup }{2024}]{yang2024cogvideox}
Zhuoyi Yang, Jiayan Teng, Wendi Zheng, Ming Ding, Shiyu Huang, Jiazheng Xu, Yuanming Yang, Wenyi Hong, Xiaohan Zhang, Guanyu Feng, et~al.
\newblock Cogvideox: Text-to-video diffusion models with an expert transformer.
\newblock {\em arXiv preprint arXiv:2408.06072}, 2024.

\bibitem[\protect\citeauthoryear{Yun \bgroup \em et al.\egroup }{2020}]{yun2020videomix}
Sangdoo Yun, Seong~Joon Oh, Byeongho Heo, Dongyoon Han, and Jinhyung Kim.
\newblock Videomix: Rethinking data augmentation for video classification.
\newblock {\em arXiv preprint arXiv:2012.03457}, 2020.

\bibitem[\protect\citeauthoryear{Zhang \bgroup \em et al.\egroup }{2019}]{zhang2019temporal}
Jiangong Zhang, Laiyun Qing, and Jun Miao.
\newblock Temporal convolutional network with complementary inner bag loss for weakly supervised anomaly detection.
\newblock In {\em ICIP}, pages 4030--4034. IEEE, 2019.

\bibitem[\protect\citeauthoryear{Zhou \bgroup \em et al.\egroup }{2023}]{zhou2023training}
Yongchao Zhou, Hshmat Sahak, and Jimmy Ba.
\newblock Training on thin air: Improve image classification with generated data.
\newblock {\em arXiv preprint arXiv:2305.15316}, 2023.

\bibitem[\protect\citeauthoryear{Zhu and Newsam}{2019}]{zhu2019motion}
Yi~Zhu and Shawn Newsam.
\newblock Motion-aware feature for improved video anomaly detection.
\newblock {\em arXiv preprint arXiv:1907.10211}, 2019.

\bibitem[\protect\citeauthoryear{Zhu \bgroup \em et al.\egroup }{2023}]{zhu2023orthogonal}
Yan Zhu, Junbao Zhuo, Bin Ma, Jiajia Geng, Xiaoming Wei, Xiaolin Wei, and Shuhui Wang.
\newblock Orthogonal temporal interpolation for zero-shot video recognition.
\newblock In {\em ACMMM}, pages 7491--7501, 2023.

\end{thebibliography}

\end{document}